\pdfoutput=1
\documentclass[11pt]{article}

\usepackage{acl}

\usepackage{times}
\usepackage{latexsym}
\usepackage{xspace}
\usepackage{amsfonts}

\usepackage[T1]{fontenc}
\usepackage[utf8]{inputenc}

\usepackage{microtype}

\usepackage{booktabs}
\usepackage{graphicx}
\usepackage{amsmath}
\usepackage{multirow}
\usepackage{multicol}
\usepackage{caption}
\usepackage{xcolor}
\usepackage{makecell}
\usepackage{arydshln}
\usepackage{tabu}
\usepackage{algorithm}
\usepackage{algorithmic}

\newcommand{\ours}{ATC\xspace}

\newcommand{\squishlist}{
	\begin{list}{$\bullet$}{
		\setlength{\itemsep}{0pt}
		\setlength{\parsep}{3pt}
		\setlength{\topsep}{3pt}
		\setlength{\partopsep}{0pt}
		\setlength{\leftmargin}{1.0em}
		\setlength{\labelwidth}{1em}
		\setlength{\labelsep}{0.5em}
   }
}
\newcommand{\squishend}{
	\end{list}
}

\title{Collaborating Heterogeneous Natural Language Processing Tasks \\ via Federated Learning}

\author{Chenhe Dong$^{1,}$\footnotemark[1]$^{\;\, ,}$\footnotemark[2] \quad Yuexiang Xie$^{2,}$\footnotemark[1]  \quad Bolin Ding$^2$ \quad Ying Shen$^{1,}$\footnotemark[3] \quad Yaliang Li$^{2,}$\footnotemark[3]\\
$^{1}$Sun Yat-sen University \quad $^{2}$Alibaba Group \\ 
dongchh@mail2.sysu.edu.cn \quad
\{yuexiang.xyx, bolin.ding, yaliang.li\}@alibaba-inc.com \\
sheny76@mail.sysu.edu.cn \\
}

\begin{document}
\maketitle
\renewcommand*{\thefootnote}{\fnsymbol{footnote}}
\footnotetext[1]{Equal contribution.}
\footnotetext[2]{Work done at Alibaba.}
\footnotetext[3]{Corresponding authors.}
\renewcommand*{\thefootnote}{\arabic{footnote}}

\begin{abstract}
The increasing privacy concerns on personal private text data promote the development of federated learning (FL) in recent years. However, the existing studies on applying FL in NLP are not suitable to coordinate participants with heterogeneous or private learning objectives. In this study, we further broaden the application scope of FL in NLP by proposing an \textsc{Assign-Then-Contrast} (denoted as \ours) framework, which enables clients with heterogeneous NLP tasks to construct an FL course and learn useful knowledge from each other. Specifically, the clients are suggested to first perform local training with the unified tasks assigned by the server rather than using their own learning objectives, which is called the {\sc Assign} training stage. After that, in the {\sc Contrast} training stage, clients train with different local learning objectives and exchange knowledge with other clients who contribute consistent and useful model updates. We conduct extensive experiments on six widely-used datasets covering both Natural Language Understanding (NLU) and Natural Language Generation (NLG) tasks, and the proposed \ours framework achieves significant improvements compared with various baseline methods. The source code is available at \url{https://github.com/alibaba/FederatedScope/tree/master/federatedscope/nlp/hetero_tasks}.
\end{abstract}

\section{Introduction}
\label{sec:intro}

Learning from the enormous quantity of data is one of the critical factors for the great success of large machine learning models~\cite{devlin2019bert, lewis2020bart, raffel2020exploring, brown2020language} in a wide range of Natural Language Processing (NLP) applications.
However, the raising privacy concerns of the public and the restriction of data protection regulations (e.g., General Data Protection Regulation~\footnote{https://gdpr-info.eu}) build up barriers across data owners, and thus make it more intractable (even unallowable) to centrally collect and store private data for training models.

Motivated by such privacy protection requirements, federated learning (FL)~\cite{mcmahan2017communication-efficient, yang2019federated} has been proposed to collaboratively train models from decentralized data in a privacy-preserving manner, which has gained rapid popularity in both academia and industry.
Previous studies~\cite{hard2018federated, ge2020fedner, qin2021improving, passban2022training} on the adoption of federated learning for NLP applications mainly follow the framework suggested by \textsc{FedAvg}~\cite{mcmahan2017communication-efficient}: Towards the same learning objective, clients independently train the model based on local data, and send their model updates to a server for federated aggregation.

Adopting such an FL framework brings several limitations in real-world NLP applications. Firstly, only participants with the same learning objective can involve in an FL course for jointly training models via federated learning. Secondly, the framework might not be suitable for participants who want to keep their learning objective private due to privacy concerns or conflict of interest, since a consensus on the learning objectives should be achieved among participants beforehand within this framework. Considering the goal of federated learning is to bridge isolated data islands rather than just coordinating participants with the same learning objective, these limitations severely block the further promotion of FL in NLP applications.

\begin{figure*}
\centering
\includegraphics[width=0.88\textwidth]{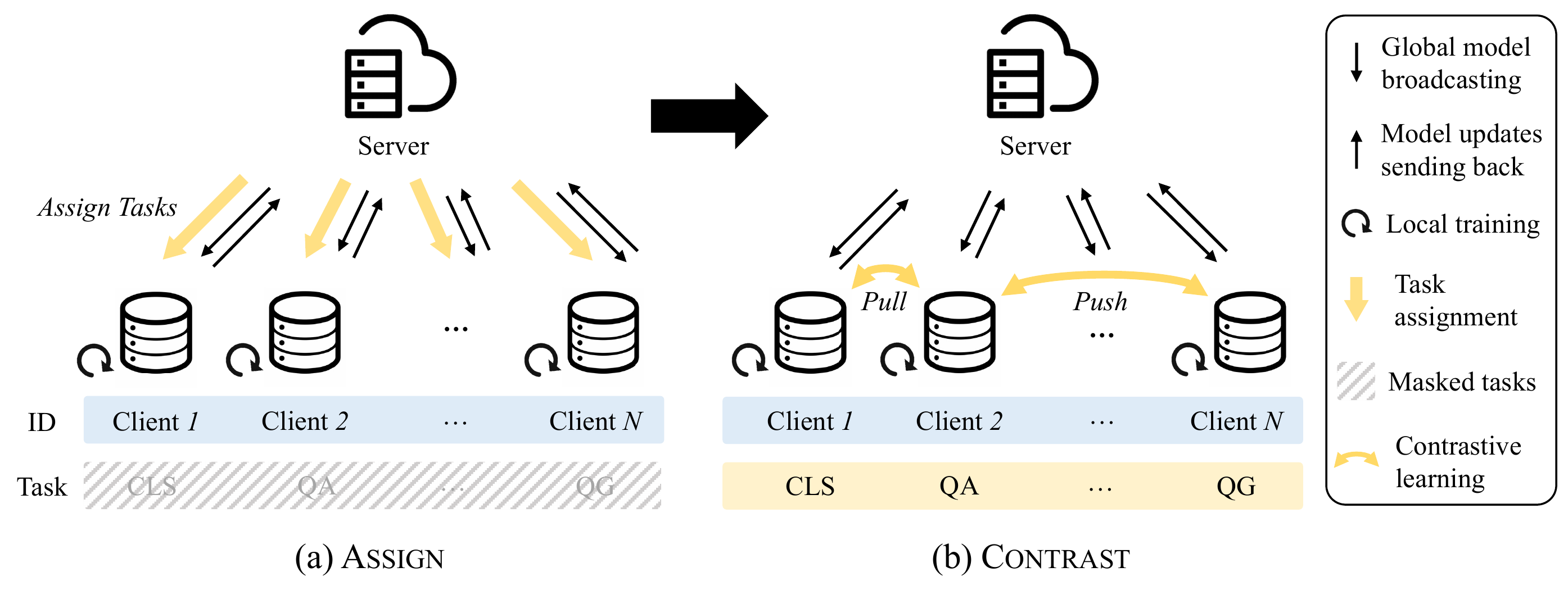}
\vspace{-0.15in}
\caption{Overview of the proposed \ours, which consists of (a) the  \textsc{Assign} training stage: Clients perform local training with the assigned tasks; and (b) the \textsc{Contrast} training stage: Clients exchange useful information with each other via optimizing a contrastive loss.\label{fig:overview}}
\end{figure*}

To tackle these limitations, in this paper, we propose a novel FL framework for NLP applications, named {\sc {\bf A}ssign-{\bf T}hen-{\bf C}ontrast} (denoted as {\bf ATC}), which enables participants with heterogeneous or private learning objectives to learn from shared knowledge via federated learning.
To be more specific, the proposed framework suggests a two-stage training paradigm in the constructed FL courses, including:
(i) {\sc Assign}. In this stage, besides broadcasting up-to-date global models, the server assigns unified tasks of local training to clients. Thus, clients can perform local training with the assigned tasks to learn knowledge from local data without using their own learning objectives.
(ii)  {\sc Contrast}. Clients perform local training according to their own learning objectives and simultaneously optimize a contrastive loss to exchange useful information with each other.
The server strategically aggregates model updates based on the measured distances among clients to make better use of these model updates.

We conduct empirical evaluations on six widely-used datasets, including various Natural Language Understanding (NLU) and Natural Language Generation (NLG) tasks such as text classification, question answering, abstractive text summarization, and question generation. 
The experimental results demonstrate the effectiveness of \ours in helping clients with heterogeneous or private learning objectives to involve and get benefit from an FL course.
Compared with several baseline methods, constructing FL courses with the proposed framework \ours achieves noticeable improvements for clients with diverse learning objectives.

\section{Preliminary}
\label{sec:preliminary}

Federated Learning (FL)~\cite{mcmahan2017communication-efficient} aims to involve multiple participants to collaboratively train global models without directly sharing their local data.
In each training round of FL, a server broadcasts the up-to-date global models to several participating clients, then clients locally train the received global model based on their local data. After local training, clients send the model updates to the server, and the server optimizes the global models by aggregating these model updates.

Formally, given that there exists one server and $N$ clients in an FL course, each client $n$ locally stores the private data $D_n$ with an amount of $|D_n|$, which won't be shared due to privacy concerns. Here we focus on horizontal federated learning where the feature spaces of clients' local data have been aligned.
The following objective function is minimized by the participants:
\begin{equation}
    F(\boldsymbol{w}) = \sum_{n=1}^{N} \frac{|D_n|}{\sum_{i=1}^{N}|D_i|}F_n(\boldsymbol{w}),
\end{equation}
where $\boldsymbol{w}$ denotes the parameters of global model, and $F_n$ denotes the objective function of client $n$.

When adopting federated learning in NLP applications, most of the previous studies~\cite{ge2020fedner, qin2021improving, passban2022training} assume that clients' learning objectives are the same. For example, \citet{passban2022training} adopts federated learning to train mixed-domain models conducting the same machine translation task.
However, in real-world applications, participants involved in an FL course might have heterogeneous or private learning objectives.

In this study, we propose to allow participants with heterogeneous or private learning objectives to construct FL courses to learn from shared knowledge via federated learning.

\section{Methodology}
In this section, we introduce the proposed \textsc{Assign-Then-Contrast} (\ours) framework, which consists of two training stages called \textsc{Assign} and \textsc{Contrast}.
An overview of the proposed \ours framework is illustrated in Figure~\ref{fig:overview}.
First of all, in Section~\S\ref{ssec:model_backbones}, we present the preparation of model backbones to enable clients with heterogeneous learning objectives to involve an FL course.
Then in Section~\S\ref{ssec:task_assign}, we describe the \textsc{Assign} training stage, the server organizes the joint learning via assigning unified tasks to clients at each training round.
In the \textsc{Contrast} training stage, clients learn useful knowledge from others with the help of a contrastive loss, as described in Section~\S\ref{ssec:contrast}.

\subsection{Model Backbones}
\label{ssec:model_backbones}
To enable clients with heterogeneous learning objectives to involve an FL course, i.e., both NLU and NLG tasks in the field of NLP,  the model backbones shared among clients are required to be aligned.
Revisiting the Transformer-based~\cite{vaswani2017attention} architectures and the standard Seq2Seq framework, the widely-used model backbone contains two fundamental components, i.e., encoder and decoder. The encoder model can be applied to NLU tasks, while the entire encoder-decoder model can be applied to NLG tasks. 

Shed lighted by such insights, we adopt the encoder-decoder architecture as the model backbone in the proposed \ours.
For the clients that only maintain the encoder for NLU tasks, 
the model backbone can be easily extended by the existing techniques such as \textsc{bert2bert}~\cite{rothe2020leveraging}, 
which initializes both encoder and decoder with the weights of BERT~\cite{devlin2019bert}. 

\begin{figure}
\centering
\includegraphics[width=0.8\columnwidth]{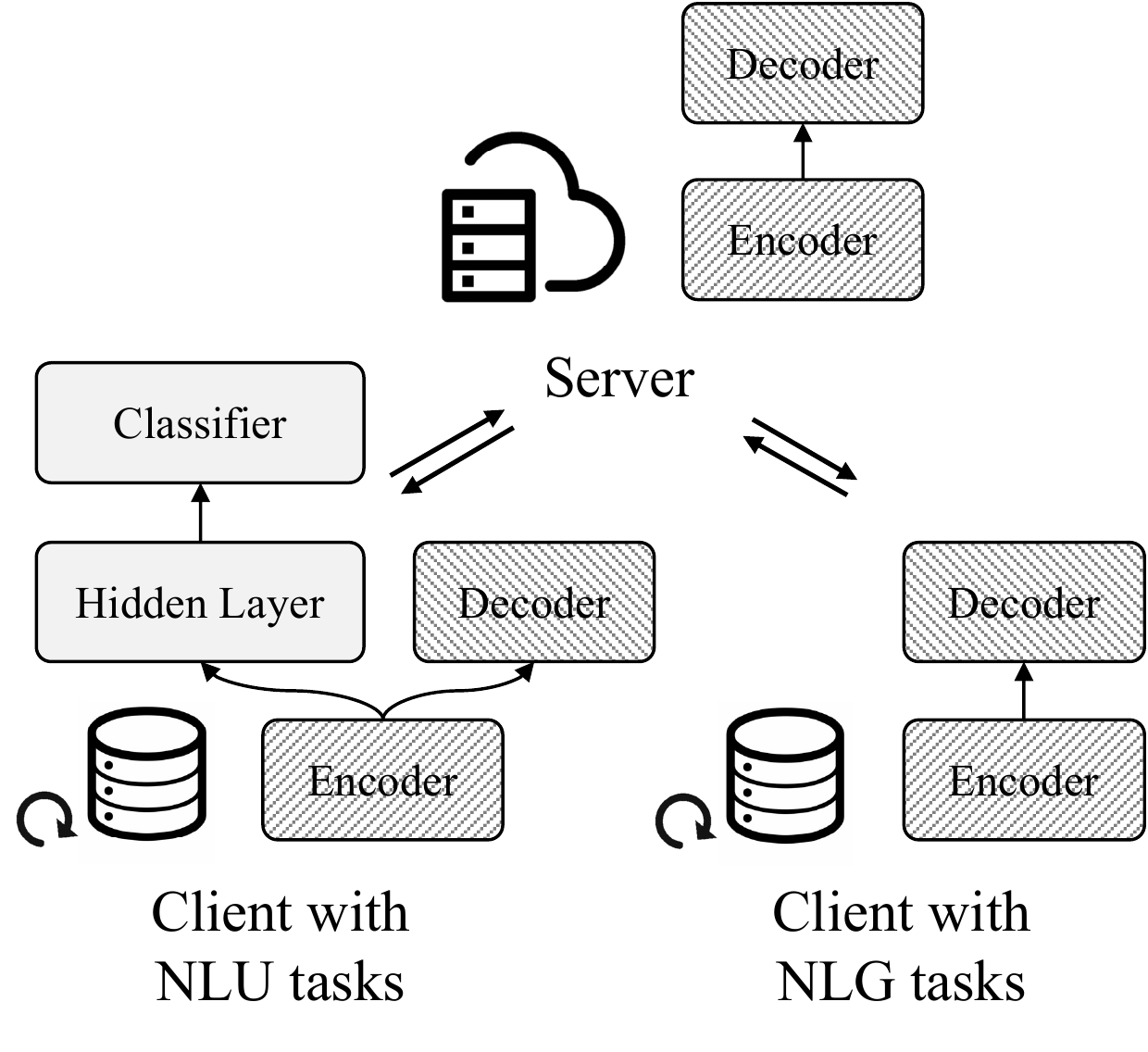}
\vspace{-0.1in}
\caption{The model backbones of participants in the proposed \ours framework.}\label{fig:backbone}
\end{figure}

In this way, the clients' model backbones that are federally learned can be aligned. Besides the model backbones, clients can maintain some private layers for personalization, e.g., the hidden layers and classifiers, as illustrated in Figure~\ref{fig:backbone}.

\subsection{\textsc{Assign}: Training with Assigned Tasks}
\label{ssec:task_assign}
The intuition behind the {\sc Assign} training stage is to enhance the exchange and integration of the general knowledge contained in clients' local data. 
The heterogeneity and inaccessibility of clients' learning objectives somehow limit such progress, therefore we design a local training approach that is agnostic to clients' learning objectives: Clients locally update the received global models based on their local data and the tasks assigned by the server, as shown in the subfigure of Figure~\ref{fig:overview}.

To be more specific, in each training round of FL, the server broadcasts the up-to-date global model and one of the prepared tasks to clients.
The prepared tasks are required to be helpful for different clients, and cannot cause privacy leakage. For example, most of the supervised learning tasks are unsuitable unless the server can annotate clients' local data without accessing them.
All these requirements inspire us to adopt the pre-training tasks in NLP, which are general and beneficial for almost all NLP downstream tasks, and, at the same time, serve in a self-supervised manner that clients can give annotations by themselves.

Here we choose two widely-used pre-training tasks as examples, including Masked Language Modeling (MLM)~\cite{devlin2019bert} and Denoising Reconstruction (DR)~\cite{lewis2020bart}.
Note that the proposed framework \ours allows more various types of tasks in the {\sc Assign} training stage.
In each training round of FL, the trainable parameters of the global model would be aggregated by the server according to the assigned tasks.
For example, when the server assigns MLM to clients, the model parameters of the encoder would be updated and aggregated; and when the assigned task is DR, the updated and aggregated parameters come from both the encoder and decoder.

In this way, participants are allowed to construct an FL course without exposing/aligning their own learning objectives and learn from shared knowledge through federal training with assigned tasks, which can also mitigate the heterogeneity in different learning objectives.

\paragraph{Extension with Clustering Algorithms}
Although the {\sc Assign} training stage tackles the heterogeneity in clients' learning objectives to some extent, the heterogeneity brought by the non-IID distributions among clients' data might lead to sub-optimal solutions of the learned global model.
Such a problem has been studied in personalized federated learning recently~\cite{li2021ditto, li2021fedbn, sattler2021clustered, xie2022personalized}, which motivates us to adopt clustering algorithms for alleviating the gradient conflicts when performing federated aggregation to further improve the proposed \ours.

Specifically, we adopt the agglomerative clustering algorithm~\cite{mullner2011modern} to distinguish clients with different domains of local data in a hierarchical manner, which is based on the cosine similarity of model updates.
Before performing federated aggregation, the server would cluster clients into several groups according to their model updates, and only the model updates of clients in the same cluster are aggregated. 
Formally, the server maintains the personalized model for each client $n$ at $t$-th training round following:
\begin{equation}
\boldsymbol{w}^{(t+1)}_{n} \leftarrow \boldsymbol{w}^{(t)}_{n} + \sum_{i \in \mathcal{C}_{n}} \frac{|D_i|}{|D_{\mathcal{C}_{n}}|} \Delta \boldsymbol{w}^{(t)}_{i},
\end{equation}
where $|D_{\mathcal{C}_{n}}|=\sum_{j\in \mathcal{C}_{n}}|D_j|$ with $\mathcal{C}_{n}$ representing the cluster that client $n$ belongs to, and $\Delta \boldsymbol{w}$ denotes the model updates, i.e., $\Delta \boldsymbol{w}^{(t)}_{n} = \boldsymbol{w}^{(t)'}_{n} - \boldsymbol{w}^{(t)}_{n}$ and $\boldsymbol{w}^{(t)'}_{n}$ implies the model after local training.

Note that rich types of personalized federated learning algorithms are compatible in the {\sc Assign} training stage similarly, and here we just introduce a representative one and focus on the proposed framework.
Further, we provide empirical results and analysis in Section~\S\ref{ssec:experimental_results} to demonstrate the effect and contribution of the adopted clustering algorithms in promoting the information exchange among similar domain corpus.

\begin{figure}
\centering
\includegraphics[width=\columnwidth]{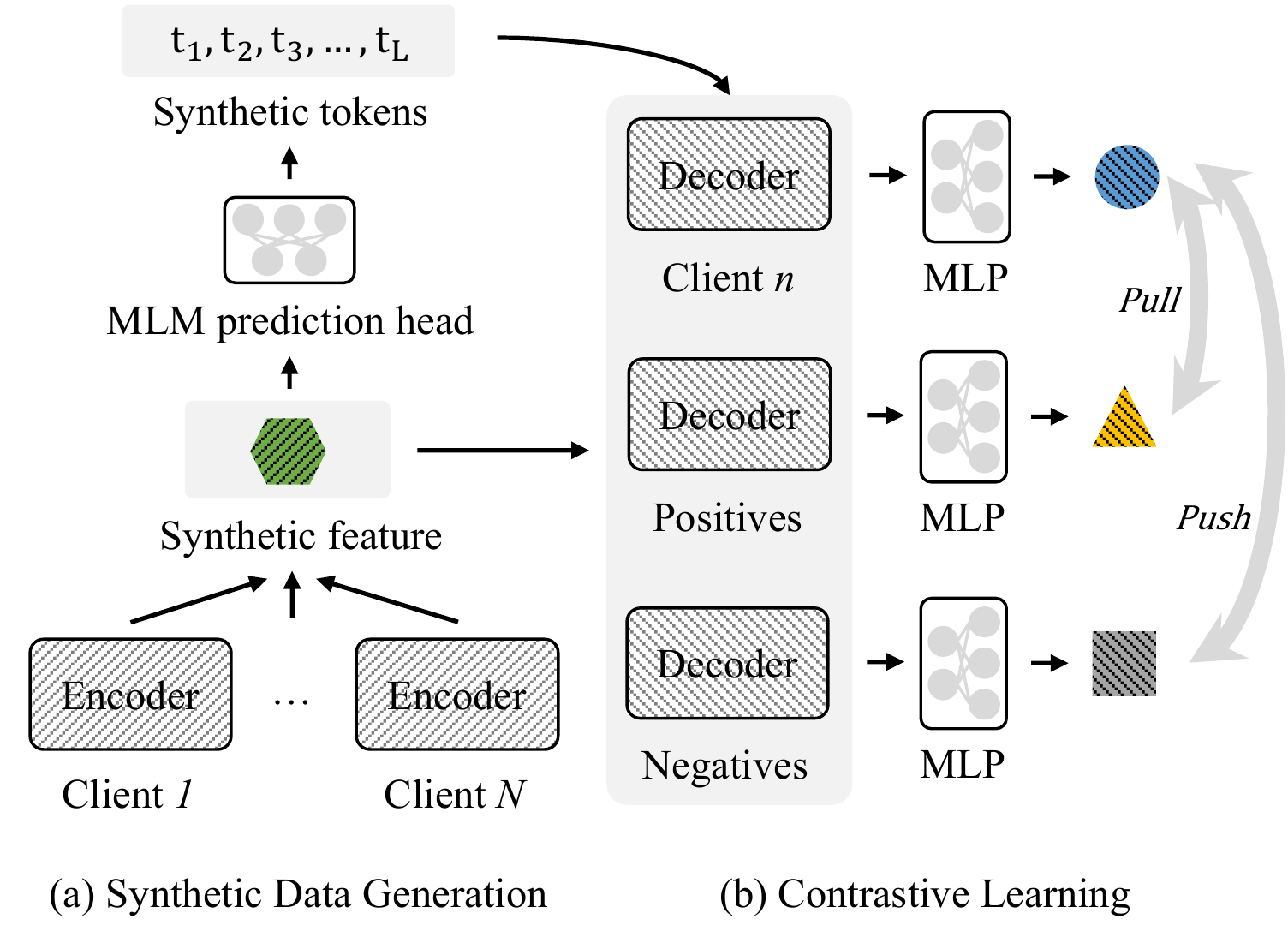}
\vspace{-0.3in}
\caption{In the \textsc{Contrast} training stage, we propose to generate a synthetic dataset (as shown in (a)) and conduct contrastive learning (as shown in (b)) for helping clients to learn useful knowledge.\label{fig:contrast}}
\end{figure}

\subsection{\textsc{Contrast}: Sharing Knowledge via Contrastive Learning}
\label{ssec:contrast}
After the {\sc Assign} training stage, clients continue to perform local training with their own learning objectives and adopt a contrastive learning technique to exchange useful information with each other, i.e., the {\sc Contrast} training stage.

Compared with applying contrastive learning in centralized training~\cite{chen2020a-simple, radford2021learning, cao2021cliff}, one of the biggest differences and challenges of applying contrastive learning in FL is to accordingly prepare or generate datasets for clients to calculate the contrastive loss without privacy leakage.

A feasible solution is using an extra public dataset, which can be fed into clients' models to generate the aligned representations or logits for computing the contrastive losses without causing new privacy leakage issues.
However, considering the non-IID distributions among clients' local data in real-world NLP applications, it is intractable to find such an eligible public dataset, as pointed out by previous studies~\cite{tan2022fedproto}.

To improve the usability of \ours framework, we propose to generate a synthetic dataset in the {\sc Contrast} training stage, as shown in Figure~\ref{fig:contrast}.
Specifically, each client randomly chooses one instance from its local data, feeds it into its local model, and takes the output of the encoder's last layer.
These outputs are sent to the server and summed up with different weights to generate a synthetic feature.
After that, the synthetic feature is passed through an MLM prediction head (reusing the MLM prediction head in {\sc Assign} training stage or loading from a pre-trained language model~\cite{devlin2019bert}) to reconstruct the corresponding tokens of this synthetic instance.
The above process is performed repeatedly to generate a synthetic dataset with appreciative quantities.
After that, the server broadcasts the synthetic dataset to all the participants.

Further, inspired by previous studies~\cite{cao2021cliff}, each client connects a multi-layer perceptron (MLP) to the decoder's last layer.
In each FL training round, clients feed the synthetic dataset into their decoders to get the averaged outputs of the MLP as the summarized representations, denoted as $\boldsymbol{h}$.
Thus, the calculation of contrastive learning loss~\cite{chen2020a-simple} can be formulated as:
\begin{align}\label{eq:cl}
\mathcal{L}_{n} &= - \log \frac{s_n^+}{s_n^+ + s_n^-}, \nonumber\\
s_n^+ &= \exp(\text{sim}(\boldsymbol{h}_n, \frac{1}{|\mathcal{C}^{+}_n|} \sum_{i \in \mathcal{C}^{+}_n} \boldsymbol{h}_i) / \tau), \nonumber\\
s_n^- &= \sum_{j \in \mathcal{C}^{-}_n} \exp(\text{sim}(\boldsymbol{h}_n, \boldsymbol{h}_j) / \tau),
\end{align}
where $\boldsymbol{h}_n$ denotes the summarized representations provided by client $n$, $\text{sim}(\cdot,\cdot)$ denotes the cosine similarity function, and $\tau$ denotes the temperature hyperparameter.
We use $\mathcal{C}_n^{+}$ and $\mathcal{C}_n^{-}$ represent the positives and negatives for client $n$ in contrastive learning, respectively.
Note that rich types of algorithms can be adopted to identify such positives and negatives.
Served as an example, here we measure the distances among clients using the cosine similarity of their model updates, and then regard one's $K$ ($K$ is a hyperparameter) closest neighbors as positives and others as negatives.

Finally, the federated aggregation in {\sc Contrast} training stage at $t$-th training round can be formulated as:
\begin{equation}
\boldsymbol{w}^{(t+1)}_n \leftarrow \boldsymbol{w}_n^{(t)} + \sum_{i \in \mathcal{C}^{+}_n} \frac{|D_i|}{|D_{\mathcal{C}^{+}_{n}}|} \Delta \boldsymbol{\tilde{w}}^{(t)}_i,
\end{equation}
where $|D_{\mathcal{C}^{+}_n}|=\sum_{j\in \mathcal{C}^{+}_{n}}|D_j|$, and $\boldsymbol{\tilde{w}}^{(t)}_n$ denotes the model updates when client $n$ performs local training to minimize both its learning objective and the contrastive loss defined in Eq.~\eqref{eq:cl}. Notably, to prevent the potential leakage of local learning objectives, each client only sends the model updates of the encoder to the server in {\sc Contrast} stage.

\section{Experiment}
\subsection{Datasets and Metrics}
We adopt six widely-used datasets for conducting a series of experiments. These adopted datasets contain various representative NLU and NLG tasks, such as text classification, question answering, text summarization, and question generation.

Specifically, we adopt two text classification datasets, i.e., IMDB~\cite{maas2011learning} and AGNews~\cite{zhang2015character-level}, which are collected from domains of review and news, respectively. 
For question answering, we adopt the SQuAD~\cite{rajpurkar2016squad} and NewsQA~\cite{trischler2017newsqa} datasets, which come from Wikipedia and news, respectively. 
Besides, we adopt a text summarization dataset CNN/DM~\cite{hermann2015teaching} and a question generation dataset MSQG~\cite{liu2021glge}, which both belong to the news domain.
Each dataset is randomly partitioned into several clients according to its quantities, as summarized in  Table~\ref{tab:datasets}, and all clients participate in one FL course.

\begin{table}[t]
\caption{Statistics of the adopted datasets.}\label{tab:datasets}
\vspace{-0.1in}
\resizebox{\columnwidth}{!}{
\begin{tabu}{lccc}
\toprule
\textbf{Dataset}& \textbf{Task}& \textbf{Domain}& \textbf{\# of Clients} \\
\midrule
IMDB& Text Classification & Review & 1 \\
AGNews& Text Classification & News & 3 \\
SQuAD& Question Answering& Wikipedia & 3 \\
NewsQA& Question Answering & News & 2 \\
CNN/DM& Text Summarization & News & 5 \\
MSQG& Question Generation & News & 4 \\
\bottomrule
\end{tabu}}
\end{table}

For evaluation metrics, we use accuracy (ACC) in text classification tasks, exact match (EM) and F1 score in question answering, and ROUGE-1/2/L (R1/2/L)~\cite{lin2004rouge}, BLEU-4 (B4)~\cite{papineni2002bleu} and METEOR (MET)~\cite{banerjee2005meteor} in text summarization and question generation following previous studies. The reported results of each dataset are the average scores of all the belonging clients.

\begin{table*}
\caption{The comparisons between the proposed \ours framework and baselines. \textbf{Bold} and \underline{underlined} indicate methods with the best and second-best performances, respectively.}\label{tab:main_res}
\vspace{-0.1in}
\centering
\setlength\tabcolsep{4pt}
\resizebox{\textwidth}{!}{
\begin{tabu}{lccccccc}

\toprule
\multirow{2}{*}{\textbf{Method}}& \textbf{IMDB}& \textbf{AGNews}& \textbf{SQuAD}& \textbf{NewsQA}& \textbf{CNN/DM}& \textbf{MSQG}& \multirow{2}{*}{\textbf{AVG}} \\
~& \textbf{(ACC)}& \textbf{(ACC)}& \textbf{(EM/F1)}& \textbf{(EM/F1)}& \textbf{(R1/R2/RL)}& \textbf{(RL/B4/MET)} \\
\midrule
\textsc{Isolated}& 78.81& 92.04& 43.25/45.14& 12.01/22.39& 37.01/15.14/33.33& 21.32/1.61/11.77& 45.38 \\
\midrule

\textsc{FedAvg}~{\small \cite{mcmahan2017communication-efficient}}& 79.14& 92.20& 46.17/48.88& 15.65/24.54& 32.64/11.35/29.59& 21.98/1.67/12.93& 45.95 \\
\textsc{FedAvg-ST}& 79.38& 92.75& 45.28/48.79& 18.68/28.47& 35.53/13.78/32.09& 23.92/1.96/13.91& 47.19 \\
\textsc{FedProx}~{\small \cite{li2020federated}}& \textbf{79.88}& 92.47& 46.28/49.13& 14.62/23.57& 27.82/7.89/25.45& 19.08/0.96/10.89& 44.97 \\
\midrule

\textsc{Ditto}~{\small \cite{li2021ditto}}& 79.48& 92.78& 42.83/46.49& 18.66/30.03& 35.41/13.77/32.02& 22.99/1.79/13.31& 46.84 \\
\textsc{FedBN}~{\small \cite{li2021fedbn}}& 79.66& 92.58& 45.32/48.38& 16.64/26.10& 32.82/11.54/29.72& 22.10/1.65/12.96& 46.23 \\
\textsc{PerCFL}~{\small \cite{sattler2021clustered}}& 77.72& 92.32& 41.76/46.60& 21.84/\textbf{34.43}& 37.21/15.28/33.51& 25.96/\underline{2.45}/15.14& 47.59 \\
\textsc{SPFL}~{\small \cite{xie2022personalized}}& 77.39& 91.91& 39.96/44.34& 20.11/32.21& 36.86/15.00/33.21& 25.23/2.30/14.68& 46.67 \\

\midrule
\textsc{ATC} (Ours)& \underline{79.72}& \underline{92.86}& \underline{46.35/49.83}& \textbf{22.58}/\underline{34.24}& \textbf{37.88/15.79/34.13}& \textbf{28.14/3.12/16.39}& \textbf{49.04} \\
\textsc{ATC} w/o \textsc{Assign}& 79.49& \textbf{92.97}& 46.05/49.48& \underline{21.98}/33.03& 37.17/15.28/33.46& 24.27/1.81/13.48& 48.26 \\
\textsc{ATC} w/o \textsc{Contrast}& 79.39& 92.71& \textbf{48.38/50.78}& 20.37/29.87& \underline{37.31/15.39/33.65}& \underline{26.40}/2.33/\underline{15.42}& \underline{48.38} \\

\bottomrule
\end{tabu}}
\end{table*}

\subsection{Baselines}
We compare the proposed \ours with the three categories of baselines, including:

\squishlist
\item \textit{Local Training}, denoted as \textbf{\textsc{Isolated}}, which implies each client independently trains its model without exchanging any information with other clients.

\item \textit{Vanilla FL and its variants}, including: 
(i) \textbf{\textsc{FedAvg}}~\cite{mcmahan2017communication-efficient}, which proposes each client locally trains the received global model based on its own data and learning objective, and sends the model updates to the server for federated aggregation. Here federated aggregation can be only performed on the sub-model that is consistently maintained by all clients;
(ii) \textbf{\textsc{FedAvg-ST}}, which constructs multiple FL courses with \textsc{FedAvg} independently. Each FL course only involves clients with the same or similar (i.e., text summarization and question generation) tasks;
(iii) \textbf{\textsc{FedProx}}~\cite{li2020federated}, which adds a proximal term to each client's loss function to reduce the instability caused by data heterogeneity in FL.

\item \textit{Personalized FL}, including: 
(i) \textbf{\textsc{Ditto}}~\cite{li2021ditto}, which trains local and global models simultaneously and fuses the local model update with the global model;
(ii) \textbf{\textsc{FedBN}}~\cite{li2021fedbn}, which suggests not sharing the batch/layer normalization parameters with others to address the non-IIDness among clients' local data;
(iii) \textbf{\textsc{PerCFL}}~\cite{sattler2021clustered}, which proposes personalized clustered FL method based on bi-partitioned clustering;
(iv) \textbf{\textsc{SPFL}}~\cite{xie2022personalized}, which defines the relationships among clients based on the similarity of their contributed gradients.

\squishend

\subsection{Implementation Details}
We use the weights of uncased BERT$_{\rm TINY}$~\cite{turc2019well-read} (with the number of layers as 2 and hidden size as 128) to initialize the encoder and decoder of the \textsc{bert2bert} model, which is provided by Huggingface~\cite{wolf2020transformers}.
For all the conducted experiments, the learning rate is set to 5e-4, the optimizer is set to AdamW~\cite{loshchilov2018decoupled} with $\beta_1 = 0.9$, $\beta_2 = 0.999$ and weight decay of 0.01. 
The linear decay of the learning rate is applied with warm-up proportion of 0.1. 

We develop the proposed \ours based on FederatedScope~\cite{federatedscope}.
In the \textsc{Assign} stage, the training round number is set to 200 and each training round contains 50 optimization steps. The batch size is set to 64, and the number of clusters is set to 5. 
In the \textsc{Contrast} stage, the training round number is set to 100 and each training round contains 200 optimization steps. 
The batch size is set to 32, the number of top-$K$ clustered clients is tuned in $[4,8,16]$, and the temperature value $\tau$ in Eq. \eqref{eq:cl} is set to 1.0.

\subsection{Experimental Results}
\label{ssec:experimental_results}
\paragraph{Comparisons}
The empirical comparisons between the proposed \ours and several baseline methods are demonstrated in Table~\ref{tab:main_res}. 
From these results, we can observe that the performances of almost all FL methods surpass that of \textsc{Isolated}, confirming that the model performance can benefit from shared knowledge via federated learning.
And the performances of \textsc{FedAvg-ST} outperform those of \textsc{FedAvg} on most of the adopted datasets, which demonstrates that applying vanilla \textsc{FedAvg} to construct an FL course among clients with heterogeneous learning objectives can lead to sub-optimal performance of the learned global model.

Further, from the comparisons between personalized FL methods and \textsc{FedAvg}, we can observe that the proposed personalized FL methods help alleviate the negative impact caused by the heterogeneity (i.e., the non-IID data distributions and various learning objectives) among clients.
However, it is worth pointing out that, all these baseline methods cannot achieve consistent improvements on the clients with various learning objectives compared to \textsc{Isolated}, which implies that not all the participants can get improvements from the shared knowledge and thus motivates us to propose the \ours framework.

The experimental results in Table~\ref{tab:main_res} demonstrate the superior performance of the proposed \ours framework in constructing FL courses among clients with heterogeneous/private learning objectives.
For example, the proposed \ours achieves consistent improvements on all the participants compared to \textsc{Isolated}, and gains an overall 3.66\% improvement.
Besides, the proposed \ours outperforms baseline methods by noticeable margins on most of the adopted datasets, especially on question answering (e.g., 0.74\% of improvements evaluated by EM on NewsQA compared to the best baseline method) and text generation tasks (e.g., 2.18\% improvement evaluated by RL on MSQG compared to the best baseline method).
These experimental results confirm the effectiveness of \ours in helping heterogeneous clients to involve and get benefit from an FL course.

\begin{table}[t]
\centering
\caption{Experimental results of ablation study.\label{tab:ablation}}
\vspace{-0.1in}
\resizebox{0.95\columnwidth}{!}{
\begin{tabu}{lcccc}
\toprule
\textbf{Method}& \textbf{CLS}& \textbf{QA}& \textbf{NLG}& \textbf{AVG} \\
\midrule
\textsc{Vanilla}& 85.67& 33.81& 18.36& 45.95 \\
\ + MLM& 85.74& 35.31& 20.97& 47.34 \\
\ + DR& 85.88& 36.71& 21.85& 48.15 \\
\ + MLM \& DR& 86.00& 36.70& \textbf{21.92}& 48.20 \\
\ + MLM \& DR \& Clu. & \textbf{86.05}& \textbf{37.35}& 21.75& \textbf{48.38} \\

\bottomrule
\end{tabu}}
\end{table}

\paragraph{Ablation Study}
In this section, we conduct a comprehensive ablation study to show the contributions of {\sc Assign} and {\sc contrast} in the proposed \ours framework. 
First of all, we skip the \textsc{Assign} training stage (denoted as ``ATC w/o {\sc Assign}'') or replace \textsc{Contrast} training stage with vanilla {\sc FedAvg} (denoted as ``ATC w/o {\sc Contrast}'') in \ours to show their effects, and the experimental results are summarized at the bottom of Table~\ref{tab:main_res}.
From these results, we can observe that the overall performance of ``\ours w/o {\sc Assign}'' and ``\ours w/o {\sc Contrast}'' drops 0.78\% and 0.66\% compared to \ours, respectively, which confirms their contributions to the proposed framework.

Further, we conduct a quantitative comparison to analyze the effectiveness of the adopted pre-training tasks (i.e., MLM and DR) and the clustering algorithms (denoted as Clu.) in the {\sc Assign} training stage.
Specifically, we firstly replace the {\sc Contrast} training stage with the vanilla {\sc FedAvg} to remove the effect of {\sc Contrast} and skip the {\sc Assign} training stage in the proposed \ours, which is denoted as {\sc Vanilla}. 
Then we recover the {\sc Assign} training stage by gradually adding the assigned tasks and clustering algorithms in the {\sc Assign} training stage. The experimental results are shown in Table~\ref{tab:ablation}, from which we can observe that both of the two adopted tasks (i.e., MLM and DR) have positive effects on the performance of the proposed \ours. For example, the overall performance of \textsc{Vanilla} has been incrementally improved by 1.39\%, 2.20\%, and 2.25\% after adopting MLM, DR, and both of them, respectively.
Besides, applying the clustering algorithms gains a further 0.18\% improvement over ``+ MLM \& DR'' and 2.43\% improvement over \textsc{Vanilla}.

\subsection{Further Discussions}
We provide further discussions to better understand how the proposed \ours help heterogeneous participants learn from each other.

\begin{figure}[t]
\centering
\includegraphics[trim=0 5 5 0, width=\columnwidth, clip]{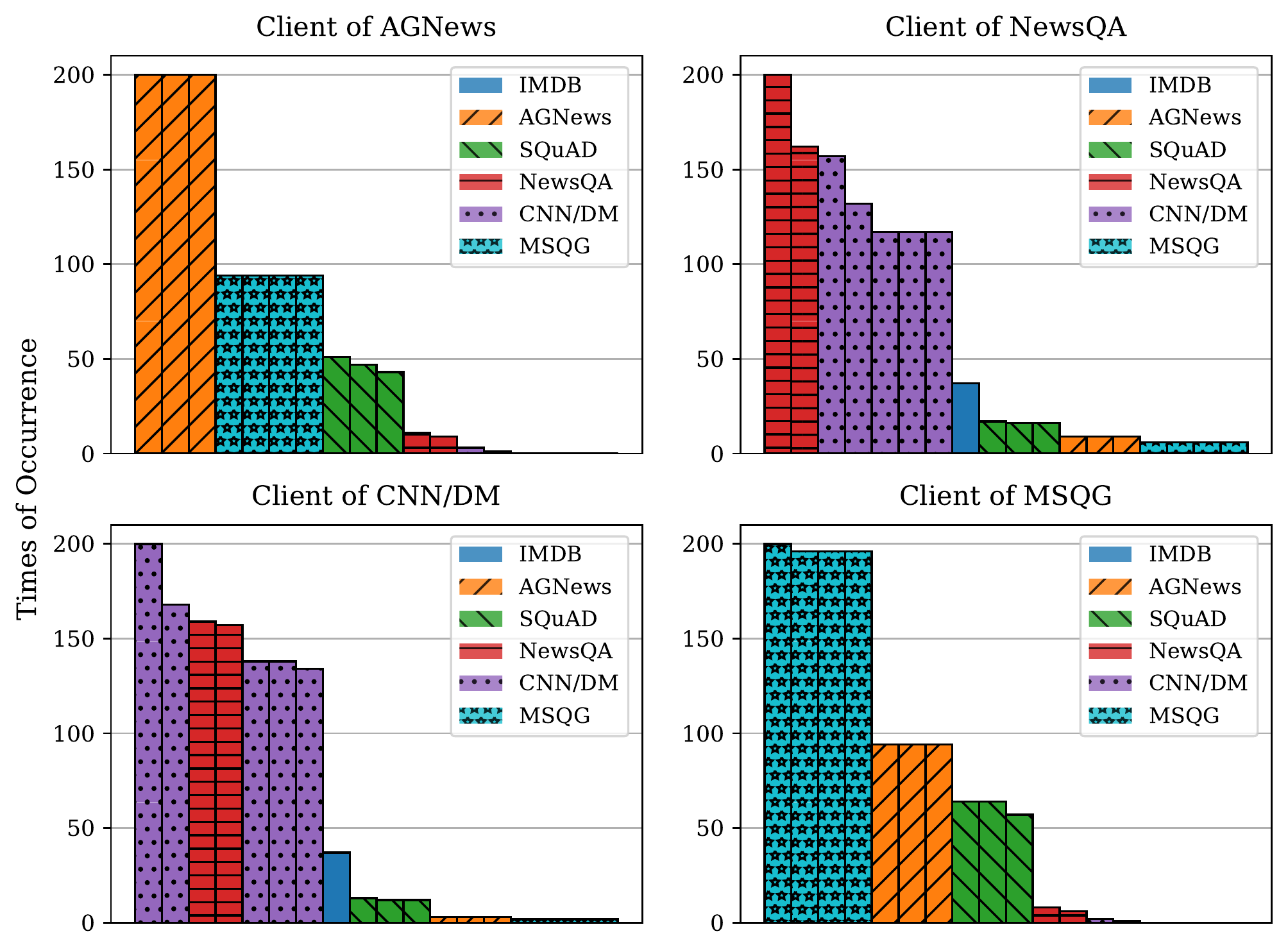}
\vspace{-0.3in}
\caption{Clustering results in \textsc{Assign}.}\label{fig:agg_assign}
\end{figure}

\paragraph{Federated Aggregation in {\sc Assign}}
Considering that clients in the same cluster would be aggregated together in the {\sc Assign} training stage, we record the statistics of clustering results and illustrated in Figure~\ref{fig:agg_assign}.
From the results, it is not surprising to observe that clients of the same datasets would be clustered into the same group and aggregated with high probability. For example, the clients of AGNews are mostly aggregated with those clients of AGNews. 
Besides, clients who have the same domain corpus are more likely to be aggregated together than those who have different domain corpus. For example, for clients of AGNews, the number of being aggregated with clients of MSQG and SQuAD is much more than that of being aggregated with others except AGNews, since AGNews, MSQG, and SQuAD are all collected from the news domain and have similar vocabulary distribution.
These results demonstrate clients learn useful knowledge from others whose corpus has similar data/vocabulary distribution in the {\sc Assign} training stage using clustering aggregation.

\begin{figure}[t]
\centering
\includegraphics[trim=0 5 5 0, width=\columnwidth, clip]{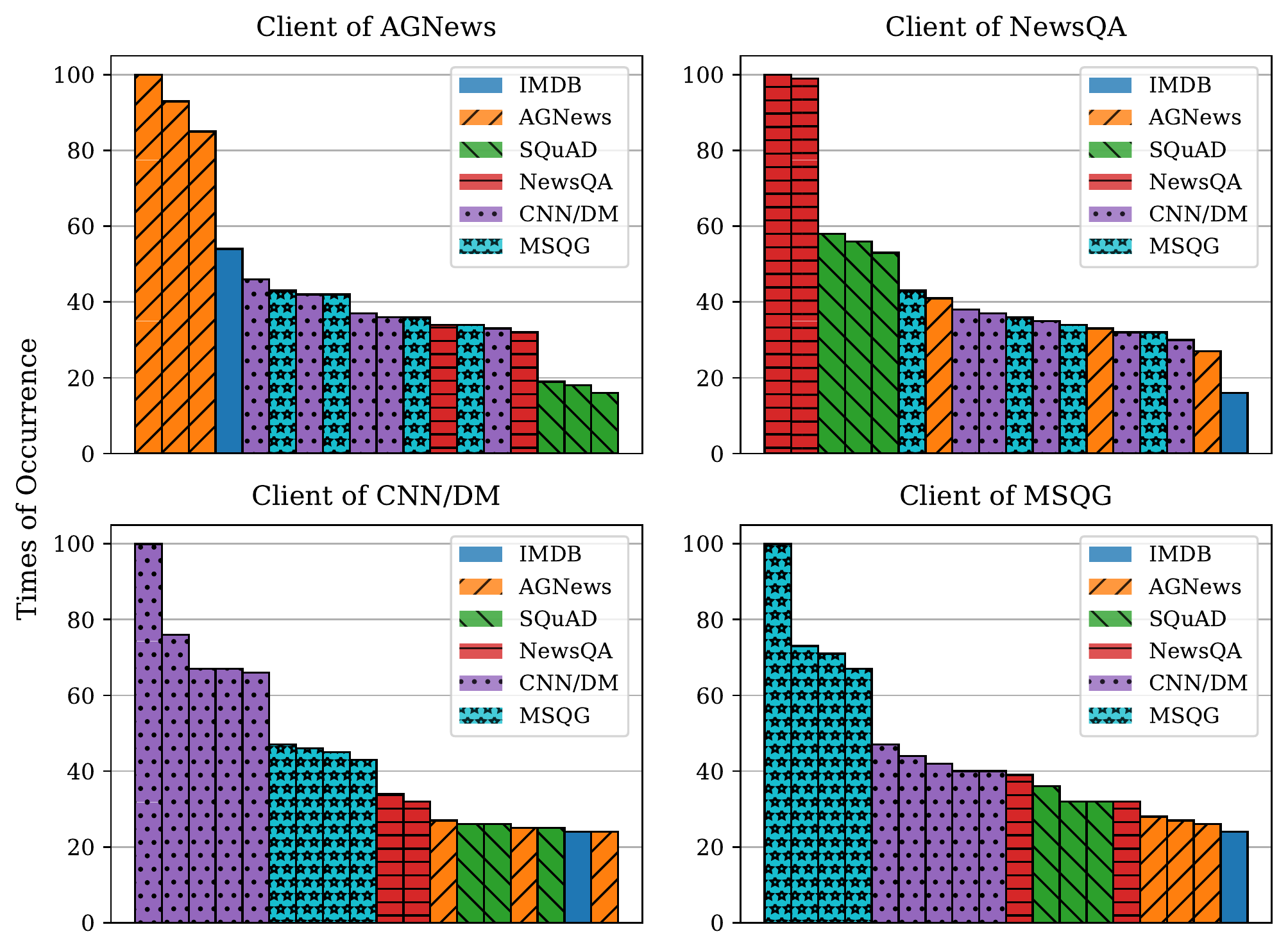}
\vspace{-0.3in}
\caption{Aggregation results in \textsc{Contrast}.}\label{fig:agg_contrast}
\end{figure}

\paragraph{Federated Aggregation in {\sc Contrast}}
We illustrate the aggregation results (i.e., the $K$ closest neighbors) in Figure~\ref{fig:agg_contrast}. From the figure, we observe that clients of the same datasets are most likely to be aggregated with each other, which is similar to the observations in {\sc Assign}.
Besides being aggregated with clients of the same datasets, clients are also aggregated with others that have similar learning objectives with high probability in the {\sc Contrast} training stage. For example, it can be observed that clients of CNN/DM are aggregated with clients of MSQG, since both CNN/DM and MSQG are text-generation tasks.
Such phenomena can be interpreted that, clients who perform local training with similar learning objectives can generate similar model updates (i.e., consistent gradient directions), therefore they would become the $K$ closest neighbors with each other.
These results further confirm that, clients can learn useful knowledge from clients whose generated model updates are similar in the {\sc Contrast} training stage of the proposed \ours.

\begin{figure}[t]
\centering
\includegraphics[trim=0 5 5 0, width=0.9\columnwidth, clip]{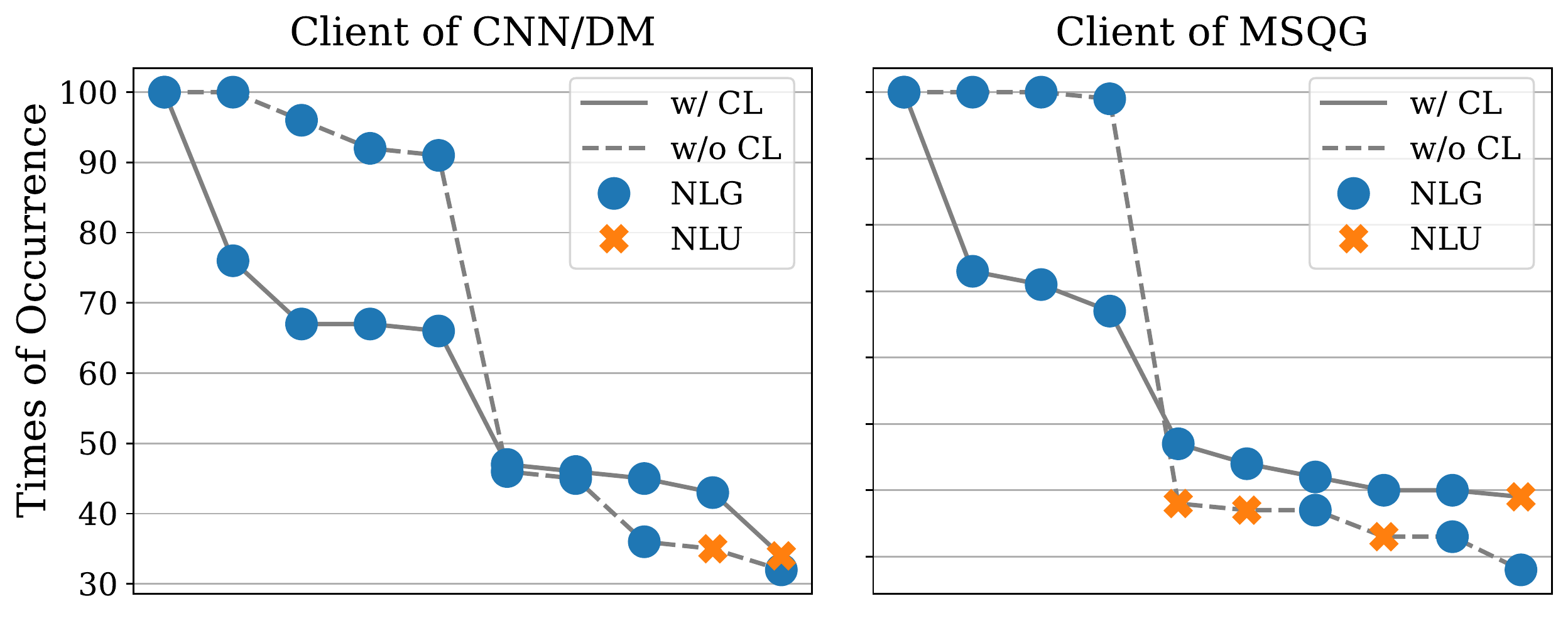}
\vspace{-0.08in}
\caption{Comparisons between applying and not applying contrastive loss in \textsc{Contrast}.}\label{fig:comp_cl}
\end{figure}

\paragraph{Effect of Contrastive Loss}
In this part, we further demonstrate the effect of contrastive loss (CL) defined in Eq.~\eqref{eq:cl} in helping clients with similar learning objectives can be aggregated with each other to learn useful knowledge in the {\sc Contrast} training stage.
Specifically, in Figure~\ref{fig:comp_cl}, we plot the neighbors of clients of CNN/DM and MSQG (both of them are NLG tasks), ranking by the distances measured by the cosine similarity of their contributed model updates. 
Comparing ``w/ CL'' with ``w/o CL'', we can observe that when applying the contrastive loss, clients with similar NLG tasks obtain a high ranking more frequently than that of ``w/o CL''. And clients with NLU tasks hardly rank in the top 10 when applying the proposed contrastive loss.
Since a higher rank implies a higher probability to be aggregated, these experimental results demonstrate that applying contrastive loss can enhance the aggregation among clients with similar learning objectives while suppressing the aggregation among clients with very different tasks.
Thus the proposed \ours can help clients with heterogeneous NLP tasks to learn useful knowledge from each other via federated learning.

\section{Related Work}
\label{sec:related_work}
Federated Learning (FL)~\cite{konevcny2016federated, mcmahan2017communication-efficient, yang2019federated} has become one of the most popular topics in both research and industrial communities in recent years.
In the community of Natural Language Processing (NLP), FL has been applied to many practical scenarios, such as keyboard prediction~\cite{hard2018federated}, named entity recognition~\cite{ge2020fedner}, semantic analysis~\cite{qin2021improving}, question answering~\cite{chen2021fedmatch}, and machine translation~\cite{passban2022training}.

Recent works in FL for NLP can be mainly divided into the following aspects: (i) Preserving data privacy. For example, \citet{sui2020feded} propose an FL framework for medical relation extraction to protect patients' private information. \citet{qin2021improving} incorporate topic memory into the FL framework to overcome the data isolation limitations. (ii) Addressing data heterogeneity. For example, \citet{chen2021fedmatch} propose a personalized backbone-patch architecture to address the non-IIDness of the question answering data. \citet{passban2022training} present a dynamic pulling FL method to efficiently train mixed-domain translation models.

Towards handling various heterogeneity of clients in federated learning, personalized federated learning~\cite{t2020personalized, fallah2020personalized, chen2022pflbench} has been widely studied to meet the demands in real-world applications, such as the non-IID distributions among clients' local data, different system resources of participants, and so on.
The existing personalization techniques in FL including regularization~\cite{dinh2020personalized, li2020federated, li2021ditto}, model mixture~\cite{mansour2020three, deng2020adaptive, li2021fedbn}, clustered learning~\cite{ghosh2020an-efficient, sattler2021clustered}, multi-task learning~\cite{smith2017federated, marfoq2021federated, xie2022personalized}, knowledge distillation~\cite{hinton2015distilling}, etc.

Different from previous studies, in this paper, we propose to enable participants with heterogeneous or private NLP tasks to involve in an FL course, so that they can exchange useful information learned from their local data with each other.

\section{Conclusion}
In this paper, we propose a novel FL framework {\sc Assign-Then-Contrast} (denoted as ATC) for helping participants with heterogeneous or private NLP tasks to learn useful knowledge from each other. The proposed framework consists of an {\sc Assign} training stage and a {\sc Contrast} training stage. In the \textsc{Assign} training stage, the server assigns tasks to clients so that clients collaboratively train the global models and share knowledge without using their own learning objectives. Then in the \textsc{Contrast} stage, clients optimize contrastive losses to learn useful information from other clients' local data. Empirical evaluations on six datasets of diverse NLP tasks demonstrate the superior performance of the proposed \ours compared to several baseline methods.
By proposing \ours, we aim to further promote the usage of FL in real-world NLP applications and inspire the community to develop new algorithms for coordinating heterogeneous participants.

\bibliography{acl}
\bibliographystyle{acl_natbib}

\end{document}